\documentclass[preprint,12pt]{elsarticle}

\usepackage{amssymb}
\usepackage{amsmath}
\usepackage{xurl}
\usepackage{booktabs}
\usepackage{ragged2e}
\usepackage{adjustbox}
\usepackage{tabularx}
\usepackage{array}
\usepackage{seqsplit}
\usepackage{hyperref}
\usepackage{graphicx}

\newcolumntype{Y}{>{\RaggedRight\arraybackslash}X}

\begin{document}

\begin{frontmatter}

\title{AI Co-Scientist for Knowledge Synthesis in Medical Contexts: A Proof of Concept} %% Article title

\author[ohri,uottawa]{Arya Rahgozar} %% Author name
\author[ohri,uottawa]{Pouria Mortezaagha}

\affiliation[ohri]{organization={Methodological Implementation Research, Ottawa Hospital Research Institute}, city={Ottawa},
            state={ON},
            country={Canada}}

\affiliation[uottawa]{organization={School of Engineering Design and Teaching Innovation, University of Ottawa}, city={Ottawa},
            state={ON},
            country={Canada}}

\begin{abstract}
\noindent
\textbf{Background:}
Research waste in biomedical science is driven by redundant studies, incomplete reporting, and limited scalability of conventional evidence synthesis workflows.

\noindent
\textbf{Objectives:}
To develop and evaluate an artificial intelligence (AI) co-scientist that enables scalable, transparent knowledge synthesis through explicit Population, Intervention, Comparator, Outcome, and Study design (PICOS) formalization.

\noindent
\textbf{Methods:}
We designed a multi-representational platform integrating relational databases, vector-based semantic retrieval, and a Neo4j knowledge graph, evaluated on dementia–sport (DS) and non-communicable disease (NCD) corpora. Automated PICOS compliance classification was performed on titles and abstracts using a Bidirectional Long Short-Term Memory (Bi-LSTM) baseline and a transformer-based multi-task classifier fine-tuned from PubMedBERT. Full-text synthesis employed retrieval-augmented generation (RAG) with hybrid vector and graph-based retrieval. Topic modeling using BERTopic identified thematic structure, redundancy, and evidence gaps. Performance was assessed via classification metrics, expert review, and RAG versus non-retrieval comparison.

\noindent
\textbf{Results:}
The transformer-based classifier achieved strong agreement with expert annotations, with study design classification accuracy of 95.7\%, while the Bi-LSTM baseline reached 87\% accuracy for PICOS compliance detection. RAG outperformed non-retrieval generation for queries requiring structured constraints, cross-study integration, and graph-based reasoning, whereas non-RAG approaches remained competitive for high-level summaries. Topic modeling revealed substantial thematic redundancy and underexplored research areas.

\noindent
\textbf{Conclusions:}
This AI co-scientist demonstrates that embedding PICOS-aware, explainable NLP into evidence synthesis workflows can improve scalability, transparency, and efficiency. The architecture is domain-agnostic and provides a practical framework for reducing research waste across biomedical disciplines.
\end{abstract}

%%Graphical abstract
% \begin{graphicalabstract}
% %\includegraphics{grabs}
% \end{graphicalabstract}

%%Research highlights
% \begin{highlights}
% \item Research highlight 1
% \item Research highlight 2
% \end{highlights}

%% Keywords

\end{frontmatter}

%% main text

\section{Introduction}
\label{sec:introduction}

Reducing research waste in knowledge synthesis is increasingly recognized as essential for efficient and high-value scientific inquiry. Research waste arises from redundant, poorly targeted, or inadequately reported studies and substantially undermines downstream evidence synthesis. Scoping reviews play a key role in mitigating these inefficiencies by systematically mapping existing evidence, identifying conceptual and methodological gaps, clarifying outcome definitions, and promoting consistent terminology, thereby helping ensure that subsequent research is necessary and appropriately directed rather than duplicative~\citep{Khalil2025ui}. Effective waste prevention further depends on robust monitoring and evaluation frameworks. Hybrid approaches that integrate survey data, quantitative indicators, and program-level assessment outperform single-method strategies~\citep{bamberger2010zt}, yet significant gaps remain in understanding which mechanisms most effectively incentivize waste reduction across research ecosystems~\citep{zacho2016ba}. Evidence suggests that durable reductions in research waste require coordinated packages of interventions rather than isolated measures, highlighting the need for stronger empirical validation~\citep{cox2010zu}.

Research waste persists across all stages of evidence generation and synthesis. Empirical studies indicate that more than 85\% of surgical randomized controlled trials (RCTs) exhibit at least one form of waste, including incomplete reporting and non-publication~\citep{chapman2019discontinuation}, with similar deficiencies observed across other biomedical domains~\citep{lu2021reporting}. These inefficiencies perpetuate delays in access to complete and reliable evidence, impairing informed decision-making by researchers, clinicians, and policymakers~\citep{dalsanto2022research,yordanov2018avoidable}.

Within evidence-based healthcare, the PICOS framework (Population, Intervention, Comparator, Outcome, Study design) remains a cornerstone for structuring research questions and guiding systematic reviews. However, its contemporary application reveals both conceptual limitations and opportunities for computational enhancement. Originally developed to support synthesis of intervention effects, PICOS has limited capacity to represent population heterogeneity, mechanistic pathways, or contextual modifiers influencing outcome variability~\citep{cumpston2020bt}. These limitations have motivated calls for synthesis approaches that integrate aggregative and configurative analyses beyond traditional meta-analytic paradigms. A further challenge is the frequent conflation of “PICO for the review” with “PICO for each synthesis,” a distinction that is insufficiently articulated in reporting guidelines and often inadequate for specifying synthesis-level eligibility and analytic criteria~\citep{cumpston2023rl}. Recent tools such as InSynQ aim to operationalize synthesis questions more explicitly, improving transparency and review planning~\citep{cumpston2023rl}. From a natural language processing (NLP) perspective, PICOS also provides a useful ontological scaffold for machine-processable representations of clinical questions, supporting automated evidence discovery, structured document processing, and downstream synthesis~\citep{Mavergames2013SystematicRA}. By imposing explicit definitions of Population, Intervention, Comparator, Outcomes, and Study design, PICOS supports efficient search strategies and eligibility criteria that directly guide downstream data extraction and synthesis, reduce retrieval of irrelevant or redundant studies, and improve comparability and reproducibility across heterogeneous evidence bases.

Methodological frameworks grounded in PICOS have been shown to improve coherence and replicability in clinical trial design~\citep{wei2024pico,zhang2024picos,lei2024picos}, yet adherence to reporting standards remains inconsistent. Guidelines such as CONSORT~\citep{davidmoher2024consort} and PRISMA~\citep{tricco2021prisma} are not uniformly applied, limiting their effectiveness in reducing bias and facilitating evidence synthesis. Authors of systematic reviews continue to report substantial challenges in managing large-scale literature searches, screening, and data extraction~\citep{santo2022research}, reinforcing the need for scalable methodological support.

These challenges have driven increasing interest in artificial intelligence (AI) methods for systematic reviews. A growing body of AI-driven NLP systems aims to accelerate literature search, screening, data extraction, and risk-of-bias assessment~\citep{OforiBoateng2024revnlp,jin2018pico}. While tools such as RobotReviewer and SciSpace demonstrate that AI can significantly reduce manual workload~\citep{robotreviewer2016tools,petri2023evidence,ge2024aisysrev}, many existing systems lack domain-specific adaptation, transparent uncertainty handling, and mechanisms for continuous updating~\citep{menezes2025gpt4medicalnotes}.

NLP methods are now integral across the knowledge synthesis pipeline. In early stages, particularly title--abstract screening, automated classifiers can rapidly filter large corpora while maintaining high recall~\citep{idnay2021hc}. At later stages, semantic retrieval and similarity-based methods enable concept-level exploration of full texts, uncovering relationships not readily captured by keyword search~\citep{zhangetal2025scientific}. Citation prioritization models further enhance efficiency and reproducibility by ranking studies by relevance~\citep{tsou2020tk}. Despite these advances, full-text processing remains a major bottleneck. Models trained primarily on titles and abstracts often degrade when applied to long-form articles due to increased linguistic complexity, heterogeneous reporting structures, and long-range dependencies~\citep{rostam2025}. Limited availability of annotated full-text corpora further constrains model development, underscoring the need for domain-adapted architectures and richer annotation resources.

The Brain--Heart Interconnectome (BHI) framework examines bidirectional interactions between cardiovascular and neurological systems, with important implications for dementia--sport (DS) research and prevention strategies~\citep{deking2021interactions,cai2024physiologic,sabor2022bhinet,catrambone2023complex}. Although the BHI literature is rapidly expanding, inefficiencies in evidence synthesis and inconsistent adherence to quality standards hinder both clinical translation and research progress. These shortcomings directly contribute to research waste, defined as redundant or low-quality research that fails to generate substantive knowledge~\citep{chalmers2009avoidable,chapman2019discontinuation,alexander2020research,rosengaard2024methods}.

In this study, we examine how NLP can be leveraged to mitigate research waste by strengthening the formalization and operationalization of PICOS across the evidence synthesis pipeline. We evaluate NLP methods for both title--abstract screening and full-text semantic analysis, focusing on how computational representations of PICOS elements can improve clarity, reduce redundancy, and enhance evidence discovery. We further assess conversational AI systems as tools to support scoping review workflows through interactive, evidence-grounded interrogation of large corpora. These methods are evaluated across two medical contexts, dementia--sport and non-communicable diseases (NCD), to examine how domain characteristics shape system performance and the potential of NLP-enabled synthesis to reduce research waste.

Situated within the broader meta-research literature that frames research waste as a systemic problem encompassing redundancy, poor reporting, irreproducibility, and misalignment with societal needs~\citep{chalmers2009avoidable}, our work builds on calls for improved reproducibility~\citep{begley2012sy} and comprehensive frameworks for adding value across the research lifecycle~\citep{westmore2023bm}. While these initiatives articulate foundational principles, they offer limited operational guidance for large-scale evidence identification and synthesis, where waste often becomes entrenched.

Motivated by this gap, we developed an AI co-scientist to support knowledge synthesis through explicit PICOS formalization, combining large-scale title--abstract classification with an interactive conversational interface. Rather than addressing research waste abstractly, we ground our investigation in the DS and NCD domains, where evidence bases are rapidly expanding and highly heterogeneous. By enabling explainable PICOS-aware screening and downstream semantic interrogation of full texts, the proposed system supports expert validation of inclusion decisions, exploration of fine-grained PICOS properties, and targeted querying of large corpora that would otherwise be infeasible to analyze manually. Our objective is to assess whether such an AI co-scientist can meaningfully augment expert judgment, reveal latent patterns of non-compliance, and inform practical assessments of research waste in specific clinical domains. Accordingly, we treat PICOS not merely as a screening heuristic, but as a unifying structural framework that constrains search, screening, extraction, and synthesis across the evidence lifecycle, providing a computational handle for identifying and mitigating research waste.

\subsection{Aim and Scope of the Proposed System}

This paper introduces an AI-driven system designed to enhance evidence synthesis in the dementia--sport and NCD domains. The proposed framework integrates:
\begin{itemize}
    \item \textbf{Automated PICOS screening:} Explainable classification to prioritize studies aligned with methodological standards.
    \item \textbf{Semantic and graph-based retrieval:} Hybrid querying using Neo4j~\citep{rafael2024neo4j} and pgVector to expose relational structure among interventions, outcomes, and populations.
    \item \textbf{Topic modeling with BERTopic:} Identification of dominant themes, redundant clusters, and underexplored research areas~\citep{grootendorst2022bertopic}.
    \item \textbf{Conversational recommender system:} Evidence-grounded responses to expert queries related to PICOS and research waste.
    \item \textbf{Interactive dashboards:} User-facing analytics integrating metadata and model-derived annotations to support exploration and policy-relevant insights.
\end{itemize}

\subsection{Overall Workflow in Knowledge Synthesis}

\begin{figure}[ht]
  \centering
  \includegraphics[scale=0.38]{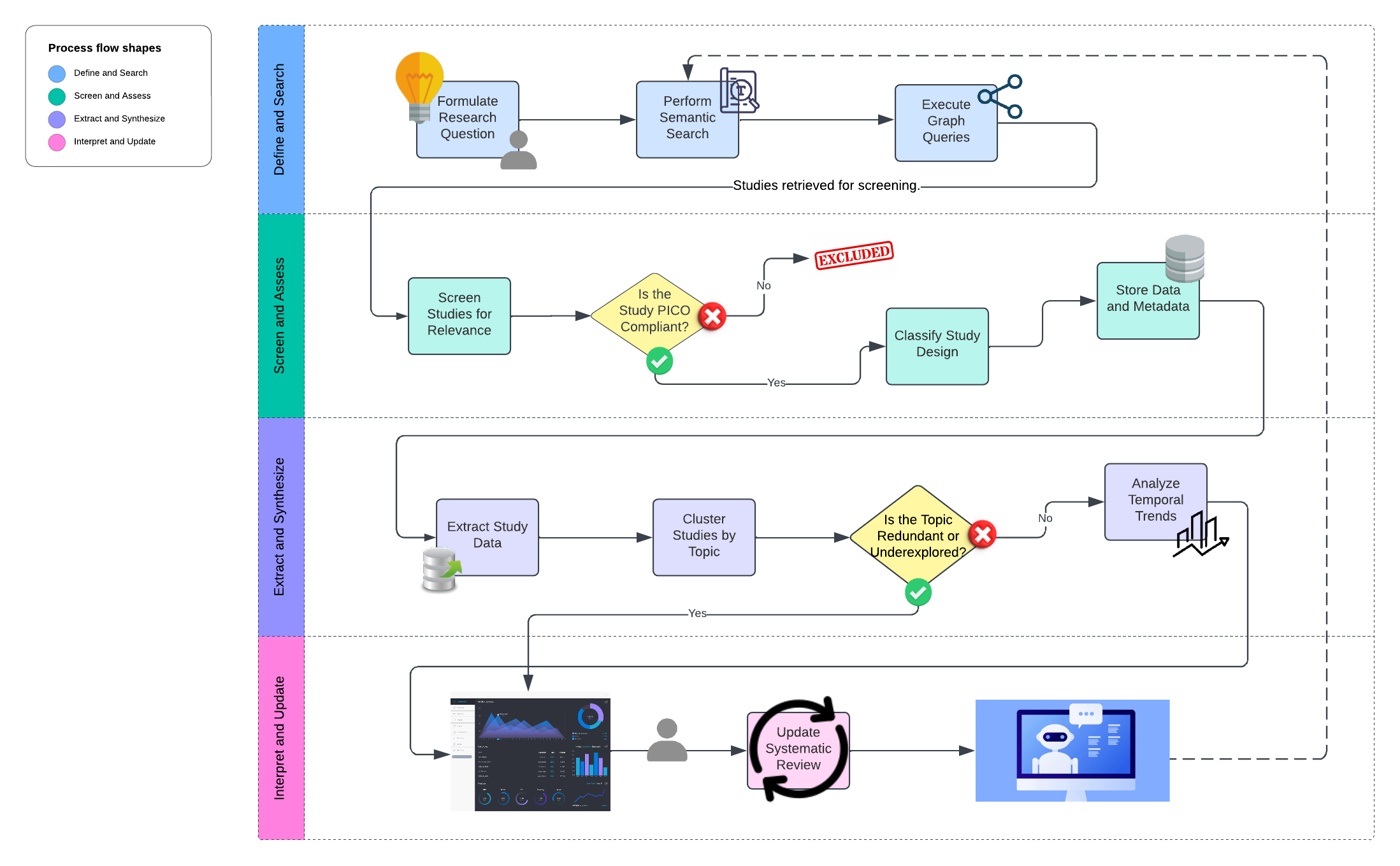}
  \caption{Workflow of the AI-driven system for knowledge synthesis in medical research.}
  \label{fig:sysrev_process_flow}
\end{figure}

Aligned with standard knowledge synthesis workflows, the system operates across four primary phases (Figure~\ref{fig:sysrev_process_flow}):
\begin{enumerate}
    \item \textit{Define and search:} PICOS-guided query formulation supported by semantic and graph-based retrieval.
    \item \textit{Screen and assess:} Automated PICOS compliance detection and hierarchical classification.
    \item \textit{Extract and synthesize:} Topic modeling and clustering to identify themes, redundancy, and emerging trends.
    \item \textit{Interpret and update:} Continuous ingestion of new literature and interactive exploration via dashboards and conversational AI.
\end{enumerate}

By prioritizing high-value evidence and minimizing redundancy, the framework aims to improve the efficiency and quality of dementia--sport and NCD evidence synthesis~\citep{legate2024semiautomated,tomczyk2024ai,gorska2024towards,ofori2024towards,toth2024automation}.

\subsection{Medical Expert and Human Requirements}
\label{sec:bhi_human_requirements}

Beyond technical considerations, system design was guided by iterative feedback from medical domain experts and stakeholders. Collaboration occurred primarily through asynchronous communication, enabling flexible engagement. Expert input informed three key areas. First, screening criteria were refined to avoid premature exclusion of studies lacking abstracts, acknowledging that incomplete records may still hold value in emerging domains. Second, eligibility criteria were aligned with evidence-based frameworks to ensure consistency in participant selection, intervention specification, comparator definition, and outcome evaluation. Third, evaluation metrics and terminology were standardized, including explicit alignment of sensitivity with recall and clarification of binary labeling conventions for study design classification.

Experts further identified the importance of capturing contextual attributes such as study setting (e.g., community-based versus institutional), leading to extensions of the hierarchical classification scheme, including explicit representation of systematic reviews as a distinct study design category. Integrating structured eligibility criteria with expert-driven feedback ensures methodological rigor while maintaining practical relevance. Continued collaboration with domain experts remains essential for adapting the system to evolving DS and NCD research landscapes.

\section{Methodology}
\label{sec:methods}

\begin{figure}[htbp]
    \centering
    \includegraphics[width=0.9\textwidth]{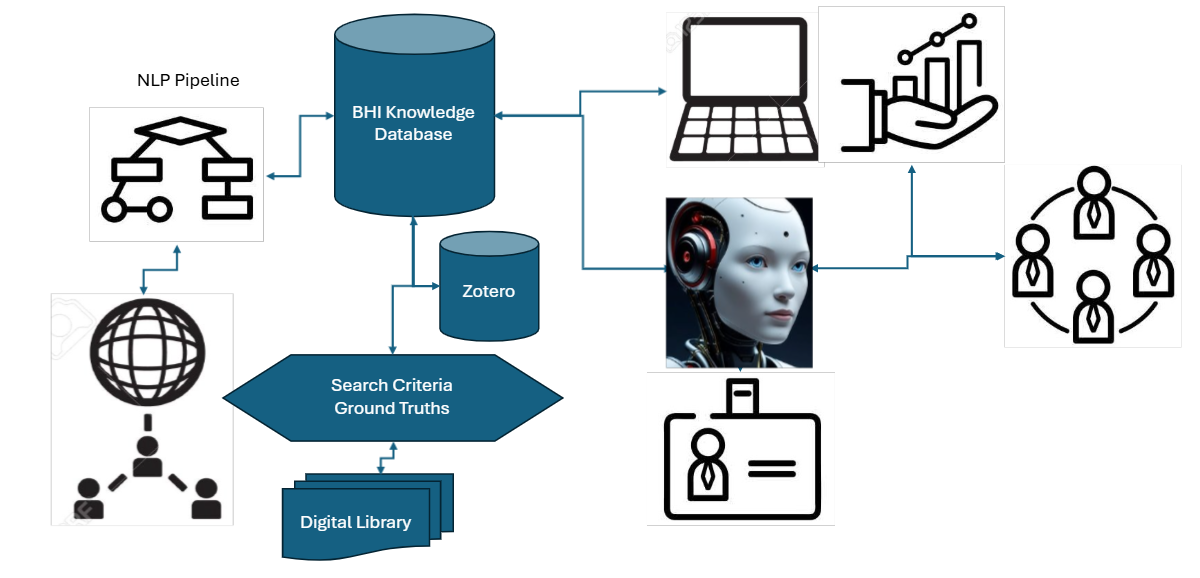}
    \caption{Overall workflow for knowledge extraction, hybrid retrieval, and expert-facing querying in the platform.}
    \label{fig:workflow}
\end{figure}

We developed an end-to-end, modular evidence synthesis platform evaluated on two corpora drawn from the dementia--sport (DS) and non-communicable disease (NCD) literature. The system supports (i) structured knowledge extraction, (ii) hybrid retrieval and evidence-grounded generation across heterogeneous data backends, and (iii) interactive, expert-facing exploration for screening, synthesis, and hypothesis generation (Figure~\ref{fig:workflow}). The architecture is intentionally \emph{multi-representational}: each corpus is maintained concurrently as (a) a normalized relational database for metadata and model outputs, (b) a passage-level vector index for semantic retrieval, and (c) an explicit knowledge graph for multi-entity biomedical reasoning.

A tool-aware orchestration layer governs multi-step reasoning, routes subtasks to appropriate storage layers, and enforces conservative grounding policies, including retrieval-first generation, evidence sufficiency checks, and mandatory citation attachment. These safeguards are designed to minimize unsupported claims and ensure traceable evidence use during synthesis. All core components are publicly available, including the agent framework, planning subsystem, graph-centric retrieval, and live database infrastructure:
\href{https://github.com/pouriamrt/medical-research-agents}{Agents},
\href{https://github.com/pouriamrt/CLARA_AI}{Planner},
\href{https://github.com/pouriamrt/POPCORN}{GraphRAG}, and
\href{https://github.com/pouriamrt/LiveDB}{LiveDB}.

The platform comprises two conceptually distinct but tightly integrated components. \emph{Kernel} refers exclusively to the domain-adapted, multi-task transformer model responsible for automated PICOS and study design classification during title--abstract screening. In contrast, the \emph{Conversational Recommender System (CRS)} denotes the expert-facing, tool-augmented interface that supports natural language querying, retrieval-augmented generation, and structured database interrogation. Kernel outputs are persisted as structured annotations and subsequently consumed by the CRS as filtering constraints and explanatory signals during evidence retrieval and synthesis.

\subsection{Corpus Preparation, Chunking, and Indexing}
\label{subsec:prep}

\paragraph{Document ingestion and normalization}
Scientific documents relevant to the Brain--Heart Interconnectome (BHI) context are ingested and normalized into a unified representation suitable for retrieval, analytics, and annotation. Bibliographic metadata, including title, authorship, venue, and publication year, are extracted at ingestion. Each document is assigned a stable identifier propagated across relational, vector, and graph layers to support reproducibility, provenance tracking, and cross-store joins.

\paragraph{Passage segmentation and chunk identifiers}
To enable retrieval at clinically and methodologically meaningful granularity, documents are segmented into discrete textual passages before embedding and indexing. Chunking reduces context dilution in long-form articles, improves retrieval precision, and supports fine-grained provenance during synthesis. Each chunk receives a unique identifier linked to its parent document, enabling passage-level citation and multi-hop graph relations (e.g., chunk-to-paper and chunk-to-author).

\paragraph{Metadata harmonization and derived annotation attachment}
Metadata fields are harmonized across ingestion sources to enable deterministic filtering, aggregation, and reproducible cohort selection. Standardized bibliographic attributes are augmented with model-derived annotations, including PICOS compliance flags and topic assignments. Persisting these annotations at the record level decouples expensive inference from interactive querying and enables consistent application of structured constraints across retrieval, dashboards, and evaluation workflows.

\subsection{Data Representation and Storage Layers}
\label{subsec:storage_layers}

\paragraph{Relational layer}
A structured relational schema stores bibliographic metadata, chunk references, and model-derived annotations, including PICOS screening outputs and topic memberships. Treating these outputs as first-class fields supports deterministic filtering, auditability, and integration with business intelligence tools for monitoring screening outcomes and longitudinal trends. This design also enables stable reuse of screening labels during retrieval and evaluation without re-running inference.

\paragraph{Vector layer}
Semantic retrieval is implemented using a PostgreSQL-backed vector store (pgVector), in which each document chunk is indexed as a dense embedding alongside rich metadata, including publication year, venue, authorship, PICOS flags, and topic assignments. This configuration supports similarity search under structured constraints and enables information needs poorly addressed by keyword search, such as cross-study comparisons of intervention--outcome relationships, subgroup-specific findings, and mechanistic descriptions expressed with heterogeneous terminology. Leveraging PostgreSQL also supports production requirements, including access control, logging, backups, and monitoring, while enabling retrieval-first RAG workflows.

\paragraph{Graph layer}
Evidence synthesis in the Dementia–Sport (DS) and non-communicable disease (NCD) domains requires reasoning over interconnected entities spanning populations, interventions, outcomes, phenotypes, study designs, venues, and authorship. We therefore represent extracted entities and relationships using a Neo4j knowledge graph, with nodes corresponding to papers, chunks, authors, topics, interventions, outcomes, and clinical descriptors. Typed edges encode relationships derived from extraction and linkage logic, enabling multi-hop neighborhood queries and pattern discovery. Compared with purely vector-based retrieval, explicit graph representations support transparent relational exploration and provide interpretable relational paths underlying synthesized answers~\citep{peng2024graphragsurvey}.

\subsection{Retrieval-Augmented Generation for DS and NCD Question Answering}
\label{subsec:rag}

\begin{figure}[ht]
    \centering
    \includegraphics[width=0.8\textwidth]{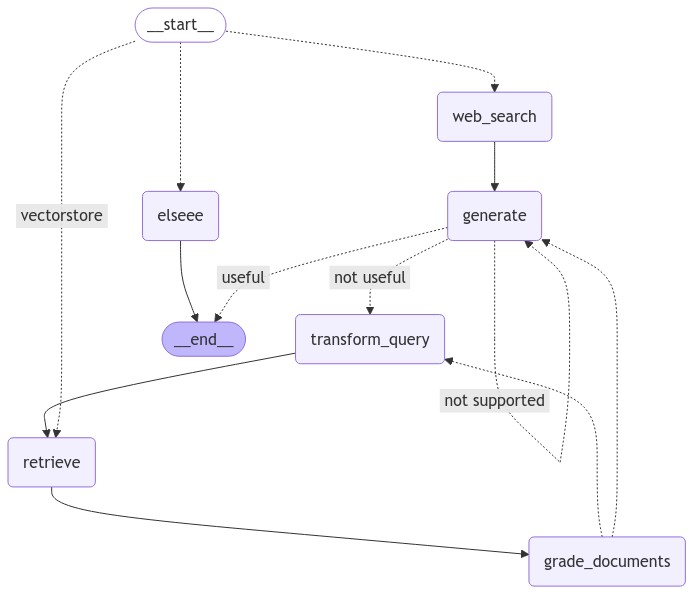}
    \caption{Workflow for query routing, hybrid retrieval, relevance grading, and corrective refinement when evidence is insufficient for grounded generation.}
    \label{fig:corrective_rag}
\end{figure}

We implement a corrective, evidence-centric retrieval-augmented generation (RAG) framework for question answering over the BHI literature. Query routing, retrieval, relevance assessment, and synthesis follow an explicit control flow that supports iterative refinement when initial evidence is insufficient (Figure~\ref{fig:corrective_rag}). Across modes, the system enforces a unified grounding requirement: all substantive claims in generated responses must be supported by retrieved passages with traceable provenance.

This requirement is operationalized through two complementary retrieval pathways: (i) graph-neighborhood retrieval using Neo4j GraphRAG for multi-entity and multi-hop reasoning, and (ii) metadata-aware semantic retrieval using a pgVector-backed store for efficient passage retrieval under structured constraints. Retrieved evidence is exposed to users and paired with sentence-level citations, supporting expert verification and reducing hallucinations.

\subsubsection{Graph-neighborhood retrieval with Neo4j GraphRAG}
\label{subsubsec:graphrag}

In the graph-centric pathway, retrieval is anchored on explicit relational structure and augmented with chunk-level text. Candidate evidence chunks are identified via hybrid retrieval over Neo4j indices using vector similarity and, when available, full-text search. Context is then expanded by traversing the local graph neighborhood (one to two hops) surrounding matched chunks. The resulting structured context preserves (i) inbound relations linking chunks to higher-level entities (e.g., papers, abstracts, authors), (ii) retrieved chunk texts, and (iii) outbound relations connecting chunks to related biomedical concepts. Edge types and directionality are retained during serialization to support both model grounding and interactive inspection. In the expert interface, graph neighborhoods can be rendered and filtered by node or edge type to enable direct examination of the relational basis underlying each response.

\subsubsection{Metadata-aware semantic retrieval with pgVector and contextual compression}
\label{subsubsec:pgvector}

In the vector-first pathway, semantic similarity search is combined with structured constraints inferred from the natural language query. A self-query retriever maps query intent to metadata predicates, including publication year, venue, authorship, PICOS flags, and executes vector search within the constrained candidate set. To reduce redundancy, maximum marginal relevance diversification is applied to the top-$k$ results. Retrieved passages are then compressed using a two-stage pipeline: an LLM-based extractor retains query-relevant spans, followed by an embedding-based filter that removes low-similarity content below a configurable threshold. The final output comprises a consolidated evidence context and a structured list of source descriptors (title, authors, venue, publication year, and screening flags), enabling auditable provenance while maintaining focused synthesis.

\subsubsection{Agentic orchestration and grounded response policy}
\label{subsubsec:orchestration}

Queries in the DS and NCD domains frequently require multi-step reasoning, including enforcement of study design constraints, cross-intervention comparison, and synthesis of methodological limitations. We therefore employ an agentic orchestration framework with explicit planning, execution, and replanning. The planner decomposes each query into a minimal sequence of atomic steps, including tool invocations (retrieval, graph traversal, or database querying) and evidence-constrained synthesis operations. The executor enforces strict tool discipline: any step requiring factual content must invoke a retrieval or database tool, and unsupported external information is prohibited.

\begin{figure}[htbp]
    \centering
    \includegraphics[width=0.4\textwidth]{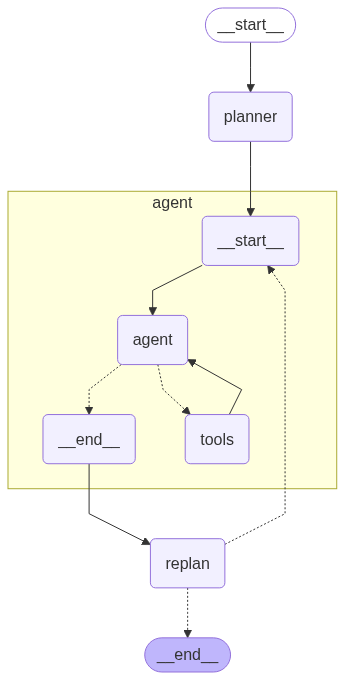}
    \caption{LangGraph-based agentic control flow for planning, iterative tool use, evidence-grounded synthesis, and replanning until sufficient evidence supports termination.}
    \label{fig:langgraph_orchestration}
\end{figure}

After each execution step, a replanner evaluates evidence sufficiency against fixed criteria: (i) coverage of all semantic constraints implied by the query, (ii) passage-level support for each substantive claim, and (iii) internal consistency across retrieved sources. The workflow terminates only when these conditions are satisfied; otherwise, targeted follow-up actions are triggered, such as additional retrieval hops or constraint refinement. This plan--execute--replan loop enforces conservative, evidence-first generation and mitigates premature synthesis (Figure~\ref{fig:langgraph_orchestration}).

When retrieval is invoked, the response policy requires sentence-level attribution linked directly to supporting passages. Retrieved sources are exposed as structured artifacts in the user interface, enabling expert verification and transparent auditing of the synthesis process.

\paragraph{Modeling choices and baselines}
We compare retrieval-augmented generation (RAG) with non-retrieval generation for BHI-specific queries. RAG is expected to improve factual accuracy and cross-study integration for queries requiring reconciliation of heterogeneous outcomes or enforcement of structured constraints, consistent with prior work~\citep{fan2024surveyragmeetingllms}. To ensure stability and reproducibility, all generation is performed using deterministic decoding (temperature $=0$).

\subsection{Automated PICOS Compliance and Explainable Classification}
\label{sec:pico_studydesign}

To accelerate screening and enable structured retrieval constraints, we implement automated classification of PICOS compliance. Predictions are persisted in the relational layer, propagated to retrieval metadata, and exposed through user interfaces to support transparent screening and query refinement. In addition to supervised classifiers, the system supports description-based title--abstract screening, in which inclusion and exclusion decisions are derived directly from expert-defined textual eligibility criteria using large language models (Appendix~\ref{app:DS_criteria}, Table~\ref{tab:appendix_ds_criteria}; Appendix~\ref{app:kernel_criteria}, Tables~\ref{tab:appendix_kernel_criteria_a} and~\ref{tab:appendix_kernel_criteria_b}).

\subsubsection{Kernel: multi-task transformer for PICOS screening}
\label{subsubsec:kernel}

Kernel is a multi-task transformer classifier fine-tuned from PubMedBERT (\path{microsoft/BiomedNLP-PubMedBERT-base-uncased-abstract}) to jointly predict Population, Intervention, Comparator, Outcome, and Study design. Each dimension is modeled as a ternary classification task (\texttt{no}, \texttt{maybe}, \texttt{yes}) to explicitly represent uncertainty and borderline cases common in abstract screening. Titles and abstracts are concatenated and encoded using a shared transformer backbone followed by parallel classification heads. Training minimizes the mean cross-entropy loss across heads to promote balanced learning across dimensions.

\paragraph{Training, calibration, and conservative self-training}
Kernel is trained using AdamW with linear warmup. Iterative expert review informs threshold calibration, label refinement, and identification of systematic failure modes, such as implicit population definitions or ambiguous intervention descriptions. To expand coverage while preserving precision, conservative self-training is applied by assigning pseudo-labels only to unlabeled instances for which all PICOS heads exceed a confidence threshold of 0.90, followed by limited retraining iterations.

\paragraph{Interpretability and auditing}
Kernel outputs per-dimension labels and confidence scores to support triage of borderline cases and targeted human review. Performance is evaluated against expert annotations using sensitivity, specificity, precision, recall, F1 score, and confusion matrices, computed per PICOS dimension and for aggregate qualification decisions used as retrieval constraints.

\subsubsection{Baseline model and study design benchmarking}
\label{subsubsec:studydesign}

For comparison under non-transformer settings, we implement a Bidirectional Long Short-Term Memory (Bi-LSTM) baseline trained on an extended PubMed-PICO dataset~\citep{jin2018pico}. For benchmarking, Kernel outputs are aligned to a binary comparator by mapping \texttt{yes} and \texttt{maybe} to \texttt{True} and \texttt{no} to \texttt{False}. In deployment, richer study design labels are retained for filtering and analytics, while the binary mapping supports comparison with legacy screening workflows.

\paragraph{Rationale for baseline selection}
The Bi-LSTM baseline reflects legacy and resource-constrained screening systems that remain prevalent in production review pipelines and provides a computationally lightweight reference point. Its role is to isolate performance gains attributable to contextualized pretraining and domain adaptation in transformer-based models such as Kernel, rather than to compete with state-of-the-art architectures.

\subsection{Live Database Connectivity and Agent-Mediated Querying}
\label{subsec:live_db}

\paragraph{Motivation}
Evidence synthesis operates in settings where corpora and annotations evolve continuously through new ingestions and revised screening decisions. To support operational use beyond static snapshots, the platform provides live connectivity to its databases, ensuring that expert queries reflect the current state of indexed evidence.

\paragraph{Live data plane}
The database infrastructure supports iterative updates across DS and NCD contexts. Literature searches are re-executed to ingest newly published or revised records, which are deduplicated, versioned, and logged at the title--abstract level to preserve provenance. The champion screening classifier is applied to refreshed corpora, producing explainable outputs that guide expert validation and prioritize full-text acquisition.

Eligible full texts are parsed, segmented, and propagated to complementary storage layers: relational databases for metadata, vector stores for semantic retrieval, and graph databases for modeling relationships across PICOS elements and study attributes. Updates are propagated immediately to downstream applications, ensuring that conversational and dashboard interfaces operate on the most current evidence base.

\paragraph{Agent--tool interface and safety constraints}
Live querying is mediated by the same tool-aware agent used for RAG workflows. For structured analytics, schema-aware SQL queries are issued to compute real-time aggregates. To reduce brittleness under schema evolution, SQL generation follows a schema-first protocol with identifier normalization and similarity-based resolution. All queries are validated before execution. Retrieval-based responses are constrained to retrieved evidence and require sentence-level attribution.

\paragraph{Session consistency and auditability}
Each interaction session is associated with a unique identifier, logging all tool calls and intermediate outputs used to construct responses. Retrieval results may be cached within a session for responsiveness, with explicit refresh available when users require the latest database state. This design balances data freshness, reproducibility, and latency.

\subsection{Topic Modeling, Visualization, and Expert Interfaces}
\label{subsec:topics_viz}

\begin{figure}[htbp]
    \centering
    \includegraphics[width=0.8\textwidth]{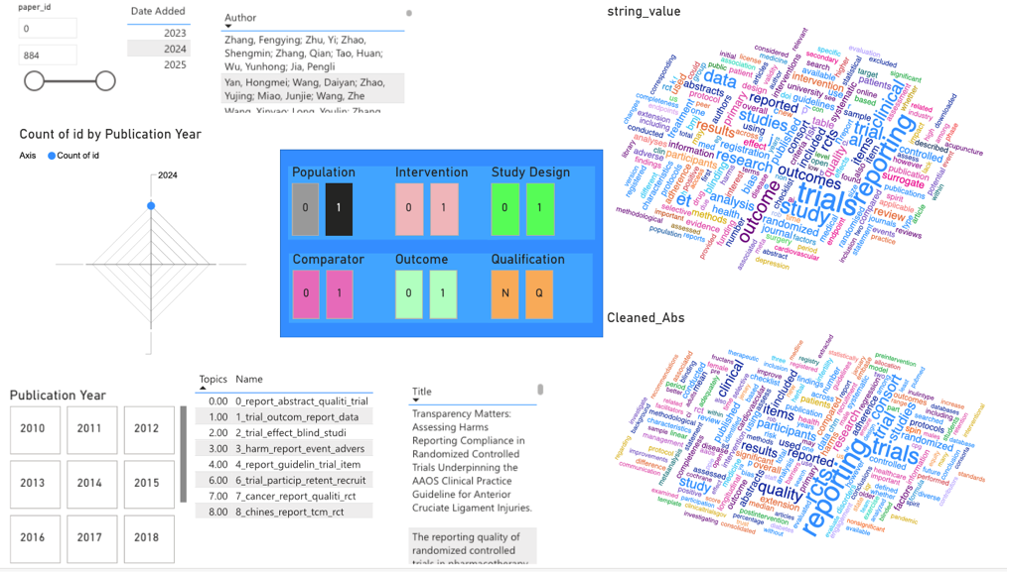}
    \caption{Power BI dashboard summarizing publication trends and screening outcomes.}
    \label{fig:dashboard}
\end{figure}

We apply BERTopic to cluster documents into coherent themes for thematic navigation, redundancy detection, and identification of evidence gaps. Topic assignments are persisted in both PostgreSQL and Neo4j, enabling longitudinal analysis and structured exploration of topic--intervention--outcome relationships. A Power BI dashboard summarizes publication trends, authorship patterns, topic evolution, and screening outcomes to support both high-level monitoring and detailed drill-down analysis (Figure~\ref{fig:dashboard}).

\subsection{Evaluation Protocol}
\label{subsec:evaluation}

The platform is evaluated along three dimensions: retrieval quality, answer faithfulness, and screening accuracy.

\paragraph{RAG versus non-RAG generation}
Domain experts assess generated responses for correctness, relevance, and completeness, with emphasis on queries requiring cross-study integration, outcome synthesis, or enforcement of study design constraints. Errors are categorized as unsupported claims, missed evidence, contradictory synthesis, or overgeneralization.

\paragraph{Retrieval performance and content fidelity}
Retrieval quality is quantified using ranking-based metrics, including Mean Reciprocal Rank (MRR). Generation fidelity is assessed using ROUGE, complemented by expert evaluation of factual consistency with retrieved passages. Redundancy among retrieved chunks and evidence coverage for key claims are also tracked.

\paragraph{Screening and study design classification}
For Kernel and baseline models, we report sensitivity, specificity, precision, recall, F1 score, and confusion matrices against expert-labeled references. Metrics are computed per PICOS dimension and for aggregate qualification decisions used operationally as retrieval constraints. Study design performance is reported under the binary mapping for benchmarking, with richer labels retained for deployment-time analytics.

\section{Results}
\label{sec:results}

We report results across four complementary components of the evaluation framework. First, we assess the performance of the explainable PICOS-based classification system applied to title--abstract screening. Second, we summarize topic modeling analyses that characterize the semantic structure of the corpora. Third, we evaluate a conversational recommender system (CRS) operating over full texts to answer expert queries related to research waste in the dementia--sport and non-communicable disease (NCD) domains. Finally, we describe an interactive dashboard that provides integrated access to curated medical databases and analytics for evidence exploration and decision support.

\subsection{PICOS Compliance Detection}

Throughout this section, the Bi-LSTM model serves as a legacy baseline representative of traditional sequence-based screening approaches, while Kernel denotes the domain-adapted transformer model developed for scalable and high-fidelity PICOS classification.

In the NCD setting, Kernel was evaluated on a gold-standard set of 284 expert-labeled inclusion and exclusion decisions. Using its ternary output space (\textit{include}, \textit{exclude}, \textit{maybe}), predictions were compared with both human annotations and a larger instruction-tuned LLM prompted with identical eligibility criteria. Kernel and the prompted LLM each achieved 95\% agreement with expert judgments, and three-way concordance among Kernel, the LLM, and the human assessor reached 91\%. Agreement beyond chance was moderate, with Cohen’s Kappa of 0.70 for Kernel versus human annotations and 0.65 for the LLM versus human annotations. Comparable or slightly higher agreement levels were observed in the Dementia--Sport corpus. Most disagreements were concentrated in abstracts with underspecified PICOS elements, where experts themselves expressed uncertainty, indicating that discordance primarily reflected borderline eligibility rather than systematic model error.

On the Dementia--Sport dataset, the Bi-LSTM baseline achieved an accuracy of 87\% for identifying PICOS-compliant abstracts. While substantially less expressive than transformer-based models, this performance illustrates the utility of automated screening for reducing downstream review burden by filtering clearly non-compliant studies early in the pipeline.

Overall, machine--human discord was largely attributable to missing or weakly articulated PICOS elements rather than definitional conflicts. In such cases, both automated models and experts tended to express uncertainty, underscoring the intrinsic difficulty of screening poorly reported abstracts. These findings indicate that Kernel closely mirrors expert reasoning in information-sparse scenarios while providing consistent, explainable screening decisions.

\subsubsection{Study Design Classification in Dementia--Sport}

We further evaluated study design classification on the Dementia--Sport corpus by benchmarking Kernel against the Cochrane Study Design classifier~\citep{thomas2021cochrane}, an SVM-based model trained on a large biomedical corpus. To ensure a fair comparison, both systems operated exclusively on titles and abstracts. Expert annotations from a gold-standard set of 100 studies served as the reference standard.

\subsubsection{Cochrane classifier performance}

Applied to the 100-study benchmark, the Cochrane classifier achieved perfect sensitivity (100\%) for identifying randomized controlled trials (RCTs) but low specificity (42\%). This imbalance resulted from 26 false positives, indicating a tendency to overclassify studies as RCTs when experts judged them to be non-randomized.

\subsubsection{Kernel performance}

Kernel was evaluated using title--abstract predictions across several domain-adapted training configurations. The deployed configuration achieved perfect sensitivity and specificity on the expert-annotated benchmark. While both Kernel and the Cochrane classifier rely on NLP-based screening, their training strategies differ substantially: the Cochrane model reflects heterogeneous biomedical data, whereas Kernel is explicitly adapted to the Dementia--Sport domain. This domain adaptation appears critical for resolving ambiguous study design cues in abstracts and reducing false-positive RCT classifications.

\subsubsection{Error analysis and comparative insights}

Detailed error analysis showed that all 26 false positives produced by the Cochrane classifier corresponded to \textit{maybe} judgments in both expert annotations and Kernel’s ternary output (\texttt{true}, \texttt{false}, \texttt{maybe}). In these borderline cases, Kernel’s explanations consistently highlighted textual cues suggestive of randomization, such as implied allocation or trial language, without explicit confirmation. The Cochrane model tended to overgeneralize these signals into binary positive predictions, whereas the Kernel preserved calibrated uncertainty. This contrast highlights the advantages of domain-specific fine-tuning and explainable ternary classification for screening nuanced clinical literature such as Dementia--Sport.

A separate validation on 164 references further assessed study design classification performance. Table~\ref{tab:studydesign_performance} summarizes both the confusion matrix counts and the derived evaluation metrics.

\begin{table}[htbp]
\centering
\renewcommand{\arraystretch}{1.25}
\setlength{\tabcolsep}{10pt}
\caption{Study design classification performance, including confusion matrix counts and derived evaluation metrics.}
\label{tab:studydesign_performance}
\begin{tabular}{@{}lcccc@{}}
\toprule
\textbf{Category} & \textbf{TP} & \textbf{FP} & \textbf{TN} & \textbf{FN} \\
\midrule
Confusion matrix & 74 & 7 & 83 & 0 \\
\midrule
\textbf{Metric} & \textbf{Precision} & \textbf{Recall} & \textbf{Specificity} & \textbf{Accuracy} \\
\midrule
Performance (\%) & 91.4 & 100.0 & 92.2 & 95.7 \\
\bottomrule
\end{tabular}
\end{table}

Together, these results demonstrate the system’s ability to rapidly prioritize methodologically rigorous studies during screening, thereby improving the quality and efficiency of downstream synthesis. Although zero-shot and instruction-tuned large language models can be applied to PICOS screening, we do not adopt them as primary baselines due to higher computational cost, reduced determinism, and limited transparency at scale. In contrast, Kernel provides stable, explainable predictions with calibrated uncertainty, making it well-suited for integration into automated and auditable evidence-synthesis pipelines.

\subsection{Topic Modeling Outcomes}
\label{sec:topic_modeling_outcomes}

BERTopic clustering revealed coherent thematic structure across the corpora, enabling identification of high-density research areas, thematic overlap, and potential evidence gaps. Beyond summary statistics and top terms, multiple visualizations were generated to examine topic prevalence and temporal evolution. An illustrative summary of all discovered topic labels and their high-frequency term representations is provided in Appendix~\ref{app:topic_modeling} (Table~\ref{tab:appendix_topic_summary}).

\begin{figure}[ht]
    \centering
    \includegraphics[width=\textwidth]{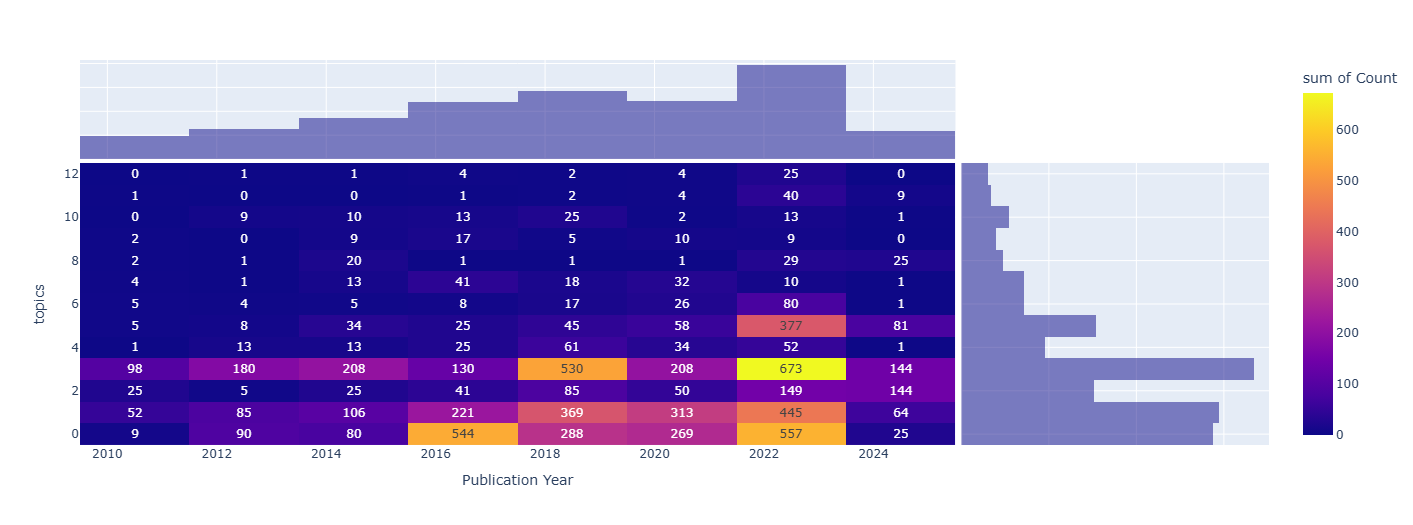}
    \caption{Heatmap of document counts by topic (y-axis) and publication year (x-axis). Higher intensity indicates greater publication density.}
    \label{fig:topic_heatmap}
\end{figure}

Figure~\ref{fig:topic_heatmap} presents a heatmap of document counts by topic and publication year (2010–2024), illustrating temporal variation in research intensity across topics. Higher color intensity denotes greater publication volume, highlighting periods of increased activity and topic-specific concentration over time.

\begin{figure}[htbp]
    \centering
    \includegraphics[width=0.75\textwidth]{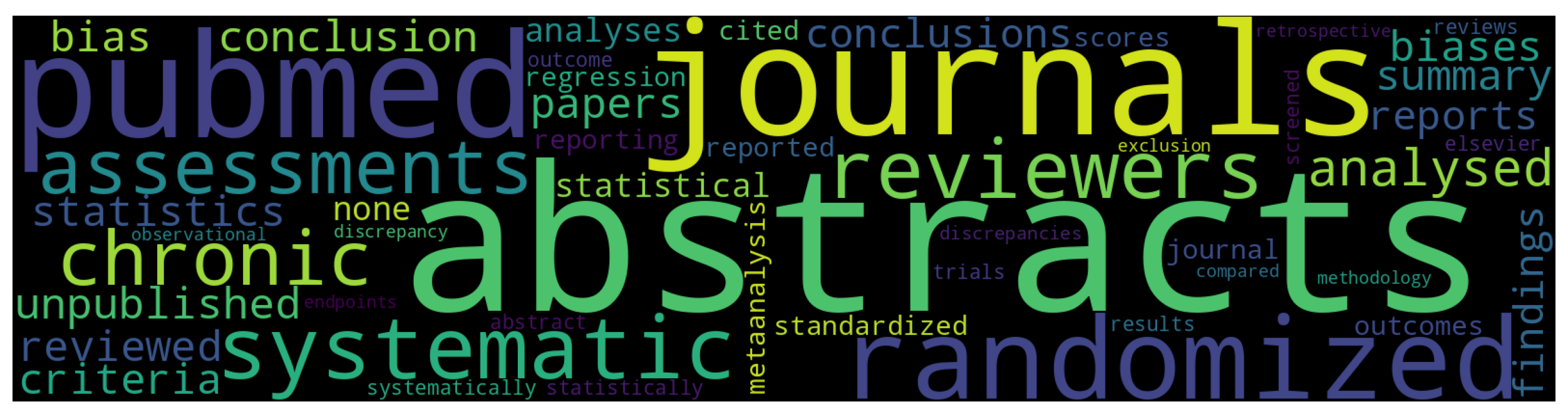}
    \caption{Representative word cloud illustrating frequently occurring terms within a topic cluster.}
    \label{fig:wordcloud_topic0}
\end{figure}

Figure~\ref{fig:wordcloud_topic0} presents a representative word cloud highlighting high-frequency terms within a selected topic. Dominant terms such as \textit{pubmed}, \textit{journals}, and \textit{randomized} point to methodological and reporting-oriented themes, providing a concise visual summary of the cluster’s semantic focus.

\paragraph{Key observations}
\begin{itemize}
    \item \textbf{Redundancy detection:} Repeated short-term cardiovascular outcome studies were identified within specific clusters, prompting closer inspection of potential redundancy and highlighting underexplored long-term neurocardiological effects.
    \item \textbf{Temporal dynamics:} Publication volume peaked between 2018 and 2022 (Figure~\ref{fig:topic_heatmap}), consistent with shifts in funding priorities and increased emphasis on standardized reporting.
    \item \textbf{Analytical granularity:} Integrating BERTopic-derived themes with relational metadata enabled fine-grained analysis of dominant interventions and outcomes within each topic.
\end{itemize}

\subsection{Performance of RAG versus Non-RAG Generation}
\label{sec:performance_rag}

Domain experts compared retrieval-augmented generation (RAG) with a non-retrieval baseline across a set of medical queries. Both configurations used the same general-purpose large language models accessed via public APIs and identical decoding settings (temperature $=0$), isolating the effect of retrieval augmentation.

Responses were evaluated along three criteria: (i) relevance of retrieved evidence, (ii) accuracy and contextual alignment of generated answers, and (iii) adequacy for informing clinical or research decision-making.

Overall, 75\% of RAG-generated responses met or exceeded expert expectations. Outcome distributions were as follows:
\begin{itemize}
    \item 30\% of queries were satisfactorily addressed by both RAG and non-RAG approaches.
    \item 25\% favored RAG, particularly for relationship-centric queries requiring integration of graph-structured evidence.
    \item 20\% favored non-RAG generation, typically for high-level thematic summaries or narrative overviews.
    \item 25\% exposed limitations requiring further optimization, primarily for queries demanding simultaneous integration of PICOS constraints, graph relationships, and topic-level context.
\end{itemize}

Topic-level representations derived from BERTopic contributed to improved query formulation and retrieval effectiveness. High-salience terms from dominant clusters were used to refine semantic constraints, enabling retrieval of more contextually relevant evidence and supporting more precise downstream synthesis.

\subsubsection{Document relevance and hallucination prevention}

The LangGraph-based orchestration effectively reduced off-target references and unsupported content by enforcing retrieval-first generation with passage-level attribution. Mapping retrieved evidence directly to user queries substantially limited hallucinations commonly observed in standalone LLM outputs, thereby improving the reliability and utility of generated responses (Figure~\ref{fig:corrective_rag}).

\subsection{Business Intelligence Dashboard}

The dashboard provides interactive access to the curated medical databases through flexible querying and filtering. Users can filter records by bibliographic metadata, topic-model-derived semantic attributes, and PICOS compliance flags generated by Kernel. Topic-derived terms function as effective surrogates for controlled vocabularies, complementing traditional indexing. This integration of structured metadata and computational annotations enables efficient exploration of evidence landscapes, identification of gaps, and informed decision-making in scoping review workflows (Figure~\ref{fig:dashboard}).

\section{Discussion}
\label{sec:discussion}

Despite substantial advances in natural language processing and clinical informatics, research waste remains pervasive due to suboptimal data curation, incomplete reporting, and manual review workflows that struggle to keep pace with rapidly expanding literatures~\citep{chalmers2009avoidable, alexander2020research,rosengaard2024methods}. The NLP pipeline and conversational recommender system (CRS) presented in this study aim to address these limitations by automating key components of knowledge synthesis while preserving expert oversight~\citep{OforiBoateng2024revnlp,toth2024automation,ge2024aisysrev,tomczyk2024ai}. Our results demonstrate a coherent alignment between core NLP tasks and the sequential stages of evidence synthesis, illustrating how automation can be integrated across the research lifecycle without compromising methodological rigor.

The workflow begins with a scalable title--abstract screening, where domain-adapted classifiers identify candidate studies from large corpora~\citep{idnay2021hc}. Only records meeting the inclusion criteria are propagated into structured downstream representations, including relational databases, knowledge graphs, and chunked vector indexes. This design ensures that subsequent analyses operate exclusively on qualified evidence, reducing noise and downstream inefficiencies. Curated evidence is then made accessible through two complementary interfaces: an interactive dashboard for structured exploration and analytics, and conversational recommender systems tailored to dementia--sport and non-communicable disease (NCD) use cases. Although optimized for these domains, the architecture is inherently generalizable and applicable to other specialties confronting large and heterogeneous evidence bases. Accordingly, the empirical findings should be interpreted as a proof-of-concept demonstrating feasibility, interpretability, and workflow integration rather than as definitive estimates of research waste prevalence.

\subsection{Automatic Classification and PICOS Compliance}

Reliable evidence synthesis depends on the selective inclusion of studies that meet PICOS criteria and adhere to robust methodological designs, such as randomized controlled trials and high-quality systematic reviews~\citep{cumpston2020bt,cumpston2023rl}. In this context, the Bi-LSTM model provided a stable baseline for abstract-level PICOS screening, enabling early exclusion of low-quality or irrelevant studies and reducing wasted effort in later review stages~\citep{jin2018pico}. Building on this baseline, our hierarchical classification strategy further prioritized methodologically rigorous designs within thematically relevant literature.

Title--abstract screening was implemented using an iterative human-in-the-loop framework in which expert judgment served as both supervisory signal and interpretive reference. Domain experts defined explicit inclusion and exclusion criteria for each PICOS dimension and provided approximately 250 gold-standard annotations. These annotations informed both fine-tuning of transformer-based models and prompt design for instruction-tuned LLMs, yielding ternary screening outputs (\textit{include}, \textit{exclude}, \textit{maybe}) that explicitly represent uncertainty.

Disagreements between model predictions and expert annotations were treated not as model failures but as informative signals for refinement. Iterative cycles of feedback, re-annotation, and validation revealed previously implicit or underspecified contextual factors, leading to progressive clarification of PICOS definitions and eligibility criteria. This calibration process continued until screening performance exceeded 95\% sensitivity and specificity relative to expert judgments, at which point the model was deemed sufficiently reliable for large-scale screening. Throughout, confidence scores and local and global explanations accompanied each prediction, supporting transparency and expert trust.

Error analysis indicated that most disagreements stemmed from nuanced semantic ambiguities rather than systematic misclassification. Explanatory signals frequently highlighted incomplete reporting of PICOS elements or studies whose primary focus fell outside the intended scope, such as methodological discussions of randomization, reporting standards, early trial termination, recruitment strategies, or trial registration practices. Other cases involved partial PICOS fulfillment, including missing comparator definitions or outcome specifications, prompting refinement of exclusion criteria through expert deliberation. Simulation-based economic models and survey studies were ultimately excluded following consultation with NCD experts, illustrating how human-machine collaboration not only improves classification accuracy but also sharpens conceptual boundaries in knowledge synthesis.

\subsection{Topic Modeling and Global Explainability}
\label{sec:addressing_redundancy}

By applying BERTopic to cluster semantically related studies (Section~\ref{sec:topic_modeling_outcomes}), the system provides a mechanism for identifying redundancy and opportunities for methodological consolidation~\citep{grootendorst2022bertopic,rosengaard2024methods}. High-frequency terms and dominant themes within clusters reveal repeated lines of investigation, enabling editors, reviewers, and funding bodies to distinguish between incremental repetition and genuinely novel contributions. In the dementia--sport and NCD domains, such signals can help redirect effort toward underexplored questions, including biomarker discovery, therapeutic stratification, and improved patient screening strategies.

\subsection{Dynamic Integration of a Live Database}

A fundamental limitation of conventional systematic reviews is the delay between publication and incorporation of new evidence. The proposed platform addresses this limitation through a continuously updated, living database architecture, ensuring that synthesis outputs reflect near real-time changes in the literature~\citep{legate2024semiautomated,gorska2024towards,toth2024automation}. This capability is particularly relevant for time-sensitive clinical and policy decisions, such as those involving acute cardiovascular conditions or neuroprotective interventions, where reliance on outdated evidence can substantially compromise downstream impact.

\subsection{Enhanced Accessibility via Conversational AI}

Conversational access to the evidence base substantially lowers barriers for non-technical stakeholders, including clinicians, allied health professionals, and policymakers\citep{Mikriukov2025pd}. By allowing users to pose domain-specific questions in natural language, the system reduces dependence on advanced data analytics expertise and narrows the gap between evidence production and application. The integration of retrieval-augmented generation (RAG) with an interactive dashboard further mitigates common limitations of standalone language models, including hallucination and weak attribution~\citep{fan2024surveyragmeetingllms}, while enabling organizations to monitor compliance trends and identify opportunities for targeted reporting guidance or policy intervention. Representative example questions illustrating the scope of PICOS-aligned queries supported by the CRS in the Dementia--Sport and NCD domains are provided in Appendix~\ref{app:crs_questions}.

Within the AI co-scientist conversational recommender system (CRS), RAG is complemented by executable tools, including a schema-aware SQL query generator. The CRS dynamically selects tools based on inferred user intent, allowing seamless transitions between semantic retrieval and structured database interrogation when appropriate.

\subsubsection{SQL Generator}

Queries implying aggregation or enumeration, such as those beginning with phrases like “how many,” consistently trigger the SQL generator, whereas descriptive or exploratory questions are routed through semantic retrieval over the document corpus. Early brittleness related to identifier case sensitivity was addressed by introducing schema-driven identifier normalization and fuzzy resolution before execution. Following these refinements, the CRS reliably answered quantitative queries, such as counts of abstracts containing specific terms, as well as analytical questions concerning topic prevalence. In each case, generated responses were directly verifiable against the underlying database, demonstrating the feasibility and reliability of tool-augmented conversational querying for structured evidence exploration.

Beyond structured queries, the CRS also supports open-ended semantic questioning over full texts, enabling clinically oriented information needs to be addressed through evidence-grounded synthesis. For example, queries regarding treatment options for advanced malignancies triggered multi-step reasoning workflows coordinated via LangGraph, integrating retrieval, synthesis, and attribution before response generation. Answers were accompanied by explicit source citations drawn from the corpus, allowing users to inspect supporting evidence and trace the reasoning process. This combination of transparency, grounded generation, and intuitive interaction supports expert confidence and facilitates interactive exploration of both conclusions and the computational processes that produced them.

\subsubsection{Waste and PICOS}

Because the conversational recommender system (CRS) has structured access to curated metadata, it can integrate bibliographic attributes such as authorship, publication date, and PICOS compliance flags with higher-level semantic reasoning over full-text content~\citep{jin2018pico,ghosh2024alpapico,dhrangadhariya2024pico,lei2024picos}. This capability enables the CRS to answer queries that combine conceptual questions about PICOS with formal research standards. For example, when asked to contrast PICO compliance with CONSORT and SPIRIT, the system correctly articulated their complementary roles: PICO structures research questions, SPIRIT governs trial protocol design and documentation, and CONSORT focuses on reporting completed trials~\citep{consort2010,spirit2013,davidmoher2024consort}. In another multi-stage query, the CRS identified and summarized the population of a specific neuro-oncology RCT corpus as patients enrolled in non-surgical randomized trials conducted between 2005 and 2014, contextualizing this information within a PICOS framework and attributing it to the appropriate sources. Across such interactions, explicit source attributions accompanied evidence-grounded responses, supporting transparency and expert verification.

The CRS also demonstrated robust performance on concept-level questions related to research waste. Queries addressing definitions, causes, ethical implications, and downstream harms of research waste yielded explanations consistent with established meta-research and research governance literature. Crucially, these responses were not generated in isolation; each was accompanied by source citations drawn from the indexed corpus, allowing users to trace claims back to the primary literature. When asked how researchers can determine whether a question has already been answered, the CRS articulated evidence-based strategies such as consulting systematic and scoping reviews, examining trial registries, and assessing overlap in PICOS elements~\citep{tricco2021prisma,Khalil2025ui,cumpston2023rl,cumpston2020bt}, again grounding recommendations in retrieved sources. Together, these results indicate that the CRS can function as a reliable AI co-scientist for reflective inquiry into research waste and responsible research practice.

\subsubsection{Aggregated Semantics with Reasoning}

A key limitation identified in the current CRS arises when queries require both semantic interpretation and explicit quantitative reasoning over metadata. In one instance, the CRS correctly addressed a conceptual question linking publication delay to null or unfavorable findings, consistent with prior meta-research. However, follow-up computational queries asking whether publication delays exceeded specific thresholds were less reliable, particularly when negative results were returned without consistent source attribution. This limitation highlights an incomplete integration between semantic reasoning, temporal metadata extraction, and verifiable computation.

To mitigate this issue, we introduce an explicit \emph{Negative Grounding} protocol for null results. When no records satisfy a query’s constraints, the CRS returns a bounded negative response together with the exact search parameters applied, including corpus scope, temporal window, study design filters, and metadata fields queried. Treating null results as first-class, auditable outputs allows users to distinguish true absence of evidence from uncertainty, incomplete ingestion, or reasoning error.

Finally, the CRS is best understood as an assistive AI co-scientist rather than an autonomous research analyst. Its effectiveness depends on alignment between user expectations and the system’s optimized competency space. Light user orientation improves performance, particularly for aggregation queries that benefit from explicit metadata framing or multi-stage decomposition. Practical constraints were also observed: concise attribution lists are most effective for disease-specific queries, with iterative follow-up supporting deeper exploration. Some requests, such as analyses over intentionally un-ingested references or complex qualitative synthesis, fall outside the current scope. Nevertheless, when engaged through iterative dialogue, the CRS demonstrated strong performance on semantically complex inquiries related to PICOS formalization and research waste, reinforcing its role as a collaborative, human-in-the-loop support tool for knowledge synthesis rather than a substitute for expert judgment.

\subsection{Business Intelligence Insights}

Dashboard-enabled filtering by individual PICOS elements revealed substantial structural gaps in the Dementia--Sport literature, providing an interpretable view of how research waste manifests at the levels of study design and reporting. By enabling isolation and quantification of compliance with each PICOS dimension, the system supports fine-grained analysis of where evidence production diverges from standards required for meaningful synthesis.

Population (P) compliance was observed in only 28\% of publications. Model explanations and expert review consistently identified insufficient population descriptions as the primary limitation, including omissions of demographic characteristics, clinical profiles, and care settings. Population reporting was frequently superficial, with limited adherence to CONSORT- or POPCORN-aligned expectations, substantially constraining interpretability and downstream aggregation.

Compliance was lowest for Intervention (I) and Comparator (C), at 15\% and 9\%, respectively. Error analysis and expert review identified these as the most semantically challenging PICOS elements to operationalize jointly in the NCD context, motivating adoption of the combined \textit{PIOS} representation for screening and downstream querying. Intervention non-compliance primarily reflected insufficient methodological specification, including missing details on treatment type, dosage, duration, or delivery, while Comparator definitions were often implicit or absent, relying on conceptual discussion rather than explicit contrasts with placebo, standard care, or alternative interventions.

Outcome (O) compliance was comparatively higher at 47\%, but remained limited by insufficient clarity. Although outcome-related concepts were often mentioned, fewer studies specified primary or secondary outcomes or described assessment procedures. Non-compliance commonly reflected proxy discussions, such as measurement frameworks or bias considerations, rather than explicit reporting of clinical endpoints.

Study Design (S) exhibited the highest compliance rate at 89\%. Nonetheless, deficiencies persisted, including references to randomized or controlled designs without explicit classification (e.g., RCT, cohort), and publications focused on reporting guidelines, statistical methods, or protocol development rather than original clinical investigations.

When considering joint PICOS compliance, only 4\% of the Dementia--Sport corpus satisfied all five criteria simultaneously. The dashboard’s capacity to surface non-compliance at both granular and aggregate levels directly informed inclusion and exclusion decisions, demonstrating how PICOS-aware analytics expose structural inefficiencies in the evidence base. These findings indicate that research waste arises from cumulative omissions across multiple PICOS dimensions rather than isolated reporting failures, underscoring the value of explainable NLP for transparent, data-driven knowledge synthesis.

\begin{figure}[htbp]
    \centering
    \includegraphics[width=0.8\textwidth]{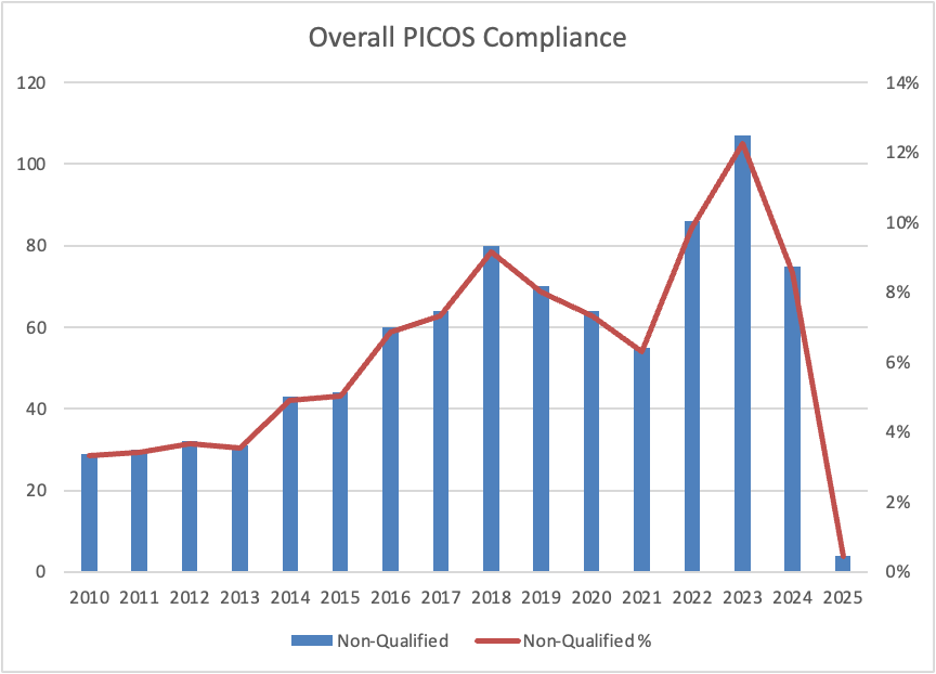}
    \caption{Trends in PICOS non-compliance and publication volume in the Dementia--Sport corpus by publication year.}
    \label{fig:PICOS_Compliance_trend}
\end{figure}

Temporal analysis further revealed a strong correlation between publication volume and PICOS non-compliance since 2010. Although data from 2024--2025 likely underestimate true trends due to ongoing publication pipelines at the time of corpus extraction, sustained increases in both publication volume and non-compliance were evident through 2023. A temporary reduction around 2019--2021 did not reflect improved reporting quality but rather parallel declines in output and non-compliance, plausibly associated with systemic disruptions during the COVID-19 pandemic (Fig.~\ref{fig:PICOS_Compliance_trend}).

Topic trends stratified by PICOS compliance illustrate how dashboard filters can interrogate reporting practices longitudinally. Non-compliant publications clustered around themes characterized by insufficient specificity in core descriptors, including population definitions, intervention details, comparator descriptions, outcome specification, and explicit study design labeling. Topic-level inspection further identified publications that appeared compliant at the abstract level but primarily addressed reporting guidelines or methodological frameworks upon closer semantic analysis, rather than reporting original clinical evidence. Aligning topic semantics with PICOS-based filtering thus enables nuanced differentiation between substantive clinical studies and meta-level discourse, enhancing transparency in inclusion decisions and identification of research waste.

\section{Limitations and Future Directions}
\label{sec:limitations}

Despite promising results, several limitations warrant consideration. First, model performance is sensitive to the quality and representativeness of training data, and underrepresented subdomains within the Brain--Heart Interconnectome may experience reduced accuracy. Second, while continuous updates improve currency, verification of newly ingested evidence requires structured validation pipelines or sustained human oversight. Third, large-scale deployment will require robust interoperability with institutional systems, raising challenges related to privacy, governance, and standardization. Finally, the platform’s effectiveness depends on user adoption; formal usability studies and iterative design refinement will be essential to maximize real-world impact.

Future work will focus on more advanced learning strategies, deeper integration with clinical and hospital informatics infrastructures, and systematic user-centered evaluations to optimize operational deployment.

\section{Ethics, Privacy, and Security}

Ethical, privacy, and security considerations were integral to the system design. All analyses were restricted to openly accessible publications, and no large language models were retrained or fine-tuned on proprietary content. Where required, generation was performed using secured local instances hosted within protected institutional computing environments, strictly for proof-of-concept evaluation. No document content was redistributed or exposed beyond controlled settings. The platform was developed solely to assess methodological feasibility, without compromising confidentiality, data governance obligations, or intellectual property rights.

\section{Conclusion}
\label{sec:conclusion}

We presented the design and evaluation of an AI co-scientist that operates alongside human experts to mitigate research waste in multiple medical contexts, with a focus on Dementia--Sport, and non-communicable diseases. Grounded in scoping review methodology, the system supports an end-to-end evidence synthesis workflow spanning title--abstract screening, full-text processing, structured extraction, and evidence-grounded synthesis, guided by explicit PICOS formalization and retrieval-first generation.

In DS and NCD evidence synthesis, PICOS serves as a practical mechanism for reducing research waste by imposing structure across the review lifecycle. Explicit definition of Population, Intervention, Comparator, Outcomes, and Study design enables efficient search strategies and eligibility criteria, reduces processing of irrelevant or redundant studies, and directly guides downstream extraction and synthesis. Consistent application of PICOS further promotes transparency, reproducibility, and cross-study comparability in heterogeneous evidence bases.

The proposed platform integrates interoperable NLP components, including ingestion and chunking pipelines, relational and graph-based representations, metadata-aware semantic retrieval, and expert-facing dashboards coupled with a conversational recommender system. Domain-adapted transformer models provide explainable PICOS classification, while live database connectivity and hybrid retrieval support auditable full-text interrogation and synthesis.

Overall, the results indicate that combining explainable screening, structured data curation, and tool-augmented conversational access can substantially reduce manual review burden while preserving expert oversight through validation and iterative refinement. Although evaluated in the Dementia--Sport and NCD domains, the architecture is domain-agnostic and readily transferable to other clinical and research settings to support scalable, transparent, and continuously updatable evidence synthesis.

\bigskip

\section*{Declarations}

\subsection*{Acknowledgement}
We gratefully acknowledge the contributions of Prof. Doug Manuel and Prof. David Moher, whose scientific leadership and guidance were instrumental throughout this work. We also thank their respective teams, including by not limited to Carol Smith, Alexandra Bodnaruc, Osmo Ramakko, and Mike Yeates, for expert annotation, rigorous testing, and constructive collaboration across all phases of the study. This research was supported by funding from the Brain-Heart Interconnectome and the Canadian Institute of Health Research.

\subsection*{CRediT authorship contribution statement}  
Arya Rahgozar and Pouria Mortezaagha designed the model, developed the computational framework, and implemented the pipeline. They both performed the experiments, statistical analyses, and wrote the manuscript. Both authors discussed the results and contributed to the final manuscript.

\subsection*{Data availability}  
The full-text articles analyzed were obtained from open-access sources. Derived datasets, intermediate outputs, and analysis scripts are available from the corresponding author upon reasonable request, subject to publisher licensing restrictions.

\subsection*{Declaration of competing interest}
The authors declare that they have no known competing financial interests or personal relationships that could have appeared to influence the work reported in this paper.

\bibliographystyle{elsarticle-harv}
\bibliography{refs}

%==============================
% \section*{Appendix}
% \clearpage
\appendix

\section{Eligibility criteria for Dementia–Sport studies}
\label{app:DS_criteria}

Table~\ref{tab:appendix_ds_criteria} summarizes the a priori eligibility criteria used to identify and screen Dementia–Sport studies, specifying inclusion requirements across participants, interventions, comparators, outcomes, and study design in accordance with the PICOS framework.

\begin{table}[ht]
\centering
\caption{Eligibility criteria for Dementia--Sport studies}
\label{tab:appendix_ds_criteria}

\renewcommand{\arraystretch}{1.15}
\small
\begin{tabularx}{\textwidth}{@{} l X @{}}
\toprule
\textbf{Domain} & \textbf{Criteria} \\
\midrule

\textbf{Type of participants} &
Older adults aged 65 years and above. Participants must be diagnosed with dementia using accepted diagnostic criteria, including DSM, ICD-10, National Institute of Neurological and Communicative Disorders and Stroke (NINCDS), Alzheimer’s Disease and Related Disorders Association (ADRDA), or CERAD-K. \\

\addlinespace
\textbf{Type of interventions} &
Exercise programs with clearly described characteristics, including type, frequency, intensity, duration, and setting. Eligible interventions included any combination of aerobic exercise, strength training, or balance training. Frequency, intensity, and duration were unrestricted. The intervention setting was not required to be explicitly described. \\

\addlinespace
\textbf{Types of comparators} &
Controlled trials in which the only systematic difference between groups was the exercise intervention. Comparator groups included usual care or social contact activities to ensure comparable levels of participant attention across groups. \\

\addlinespace
\textbf{Types of outcomes} &
\textbf{Primary outcomes:} cognition; activities of daily living; neuropsychiatric symptoms (e.g., agitation, aggression); depression; mortality.  
\newline
\textbf{Secondary outcomes:} caregiver burden; quality of life; mortality; healthcare service costs. \\

\addlinespace
\textbf{Study design} &
Individually randomized or cluster-randomized controlled trials (RCTs), including both parallel-group and cross-over designs. For cross-over trials, only data from the first intervention phase before crossover were considered.  
\newline
\textbf{Excluded designs:} systematic reviews (with or without meta-analysis), observational studies (e.g., cohort or longitudinal designs), qualitative studies, and other non-randomized study designs. \\

\bottomrule
\end{tabularx}
\end{table}

\section{Eligibility criteria for Kernel-based study screening}
\label{app:kernel_criteria}

Tables~\ref{tab:appendix_kernel_criteria_a} and~\ref{tab:appendix_kernel_criteria_b} summarize the inclusion and exclusion criteria applied by the Kernel classifier for automated screening of simulation-based population health studies.

\begin{table}[htbp]
\centering
\caption{Kernel eligibility criteria for simulation-based population health studies: study characteristics and population}
\label{tab:appendix_kernel_criteria_a}

\renewcommand{\arraystretch}{1.15}
\small
\begin{tabularx}{\textwidth}{@{} p{0.15\textwidth} p{0.4\textwidth} p{0.4\textwidth} @{}}
\toprule
\textbf{Domain} & \textbf{Include} & \textbf{Exclude} \\
\midrule

\textbf{Study characteristics} &
Studies in which computational simulation modelling is the primary analytical method used to address population health research objectives. Eligible approaches include system dynamics models; agent-based models; microsimulation and macrosimulation; discrete event simulation; Markov models and Monte Carlo simulation; life table methods for policy analysis; and computational attributable risk models. &
Observational or experimental studies without simulation modelling. Predictive modelling studies covered by TRIPOD or TRIPOD-AI, unless simulation modelling is explicitly used. Methodological or statistical studies without substantive population health questions. Meta-analyses without scenario simulation. \\

\addlinespace
\textbf{Population} &
Studies examining real or generalizable human populations or subpopulations defined by geographic, demographic, or social characteristics with relevance to population health or policy. &
Individual-level clinical studies, trial-design simulations, abstract or theoretical populations, animal or laboratory studies, education-focused simulations, or individual risk prediction tools. \\

\bottomrule
\end{tabularx}
\end{table}

\begin{table}[ht]
\centering
\caption{Kernel eligibility criteria for simulation-based population health studies: intervention and outcomes}
\label{tab:appendix_kernel_criteria_b}

\renewcommand{\arraystretch}{1.15}
\small
\begin{tabularx}{\textwidth}{@{} p{0.15\textwidth} p{0.4\textwidth} p{0.4\textwidth} @{}}
\toprule
\textbf{Domain} & \textbf{Include} & \textbf{Exclude} \\
\midrule

\textbf{Intervention / exposure} &
Simulation studies evaluating the health impact of exposures, interventions, or policies on non-communicable disease outcomes through explicit or implicit scenario comparisons. &
Methodological studies without applied health scenarios; behavioral simulations without NCD relevance; sensitivity or validation exercises without scenario comparison; economic analyses focused primarily on costs. \\

\addlinespace
\textbf{Outcome} &
Studies reporting outcomes related to WHO-defined non-communicable diseases, including mortality, life expectancy, YLL, QALYs, DALYs, disease burden, or major NCD risk factors. &
Studies focused on non-NCD conditions, intermediate physiological outcomes only, or healthcare system outcomes without explicit linkage to NCD health impacts. \\

\bottomrule
\end{tabularx}
\end{table}

\section{BERTopic Cluster Definitions and Representative Terms}
\label{app:topic_modeling}

Table~\ref{tab:appendix_topic_summary} provides an illustrative overview of the extracted topics, including topic labels, high-frequency terms, and representative document groupings.

\begin{table}[ht]
\centering
\caption{Illustrative BERTopic output for all discovered topics, showing topic label (Name) and high-frequency terms (Representation). Topic $-1$ corresponds to outlier documents not assigned to a coherent cluster.}
\label{tab:appendix_topic_summary}

\setlength{\tabcolsep}{5pt}
\renewcommand{\arraystretch}{1.15}
\small
\begin{adjustbox}{max width=\textwidth}
\begin{tabularx}{\textwidth}{@{} r Y Y @{}}
\toprule
\textbf{Topic} & \textbf{Name} & \textbf{Representation} \\
\midrule
0  & \textsc{\seqsplit{report\_trial\_studi\_rct}} &
\textit{\seqsplit{\{report, trial, studi, rct, data, use, \dots\}}} \\
1  & \textsc{\seqsplit{outcom\_trial\_regist\_registr}} &
\textit{\seqsplit{\{outcom, trial, regist, registr, primari, \dots\}}} \\
2  & \textsc{\seqsplit{blind\_trial\_bia\_effect}} &
\textit{\seqsplit{\{blind, trial, bia, effect, outcom, \dots\}}} \\
3  & \textsc{\seqsplit{harm\_report\_trial\_event}} &
\textit{\seqsplit{\{harm, report, trial, event, advers, \dots\}}} \\
4  & \textsc{\seqsplit{qualiti\_rct\_report\_abstract}} &
\textit{\seqsplit{\{qualiti, rct, report, abstract, method, \dots\}}} \\
5  & \textsc{\seqsplit{size\_sampl\_sampl size\_calcul}} &
\textit{\seqsplit{\{size, sampl, sampl size, calcul, power, \dots\}}} \\
6  & \textsc{\seqsplit{trial\_guidelin\_item\_report}} &
\textit{\seqsplit{\{trial, guidelin, item, report, develop, \dots\}}} \\
7  & \textsc{\seqsplit{journal\_report\_orthodont\_rct}} &
\textit{\seqsplit{\{journal, report, orthodont, rct, publish, \dots\}}} \\
8  & \textsc{\seqsplit{subgroup\_trial\_analys\_subgroup}} &
\textit{\seqsplit{\{subgroup, trial, analys, subgroup analys, \dots\}}} \\
9  & \textsc{\seqsplit{chines\_tcm\_report\_rct}} &
\textit{\seqsplit{\{chines, tcm, report, rct, qualiti, \dots\}}} \\
10 & \textsc{\seqsplit{pro\_rct\_cancer\_report}} &
\textit{\seqsplit{\{pro, rct, cancer, report, qualiti, \dots\}}} \\
11 & \textsc{\seqsplit{abstract\_report\_ci\_trial}} &
\textit{\seqsplit{\{abstract, report, ci, trial, statist, \dots\}}} \\
12 & \textsc{\seqsplit{recruit\_retent\_particip\_strategi}} &
\textit{\seqsplit{\{recruit, retent, particip, strategi, \dots\}}} \\
13 & \textsc{\seqsplit{trial\_particip\_patient\_organ}} &
\textit{\seqsplit{\{trial, particip, patient, organ, platform, \dots\}}} \\
\bottomrule
\end{tabularx}
\end{adjustbox}
\end{table}

\section{Example CRS Competency Questions Aligned with PICOS}
\label{app:crs_questions}

To preserve privacy and data governance constraints, we do not report verbatim system responses. Instead, we present representative competency questions illustrating the types of evidence-grounded, PICOS-aligned queries the Conversational Recommender System (CRS) is designed to support in the Dementia--Sport and non-communicable disease (NCD) domains.

\begin{table}[ht]
\centering
\caption{Example CRS competency questions aligned with Population (P)}
\label{tab:crs_population}
\begin{tabularx}{\textwidth}{@{}lX@{}}
\toprule
\textbf{Domain} & \textbf{Representative Questions} \\
\midrule
Dementia--Sport &
What populations are most frequently studied (e.g., age group, dementia stage, care setting)? \newline
How often are demographic characteristics explicitly reported? \newline
Which dementia subtypes are most represented in physical activity interventions? \\
\addlinespace
NCD &
Which NCD populations dominate the corpus (e.g., cardiovascular disease, diabetes, cancer)? \newline
How does population reporting vary by NCD subtype and publication year? \newline
Are vulnerable or underrepresented populations systematically included or excluded? \\
\bottomrule
\end{tabularx}
\end{table}

\begin{table}[ht]
\centering
\caption{Example CRS competency questions aligned with Intervention (I)}
\label{tab:crs_intervention}
\begin{tabularx}{\textwidth}{@{}lX@{}}
\toprule
\textbf{Domain} & \textbf{Representative Questions} \\
\midrule
Dementia--Sport &
What types of sport or physical activity interventions are most commonly evaluated? \newline
How often are intervention parameters (intensity, duration, frequency) sufficiently reported? \newline
Which interventions are associated with cognitive versus functional outcomes? \\
\addlinespace
NCD &
Which intervention classes dominate NCD research? \newline
Are intervention protocols consistently described across studies? \newline
How has intervention complexity evolved over time? \\
\bottomrule
\end{tabularx}
\end{table}

\begin{table}[ht]
\centering
\caption{Representative CRS competency questions aligned with Comparator (C)}
\label{tab:crs_comparator}
\begin{tabularx}{\textwidth}{@{}lX@{}}
\toprule
\textbf{Domain} & \textbf{Example Questions} \\
\midrule
Dementia--Sport &
How frequently do studies include an explicit comparator group? \newline
What types of comparators are used (e.g., usual care, placebo, alternative activity)? \newline
In studies lacking comparators, what methodological justifications are provided? \\
\addlinespace
NCD &
How often are standard-of-care comparators used? \newline
Are head-to-head intervention comparisons common in specific NCD subdomains? \newline
Which NCD areas exhibit the highest rates of missing comparator information? \\
\bottomrule
\end{tabularx}
\end{table}

\begin{table}[ht]
\centering
\caption{Representative CRS competency questions aligned with Outcome (O)}
\label{tab:crs_outcome}
\begin{tabularx}{\textwidth}{@{}lX@{}}
\toprule
\textbf{Domain} & \textbf{Example Questions} \\
\midrule
Dementia--Sport &
What primary outcomes are most frequently reported (e.g., cognition, mobility, quality of life)? \newline
How often are primary and secondary outcomes explicitly distinguished? \newline
Are outcome measures standardized across studies? \\
\addlinespace
NCD &
What outcome categories dominate NCD research (e.g., mortality, biomarkers, functional outcomes)? \newline
How frequently are outcomes clearly operationalized and measurable? \newline
Which NCD domains show the greatest outcome-reporting heterogeneity? \\
\bottomrule
\end{tabularx}
\end{table}

\begin{table}[ht]
\centering
\caption{Representative CRS competency questions aligned with Study Design (S)}
\label{tab:crs_studydesign}
\begin{tabularx}{\textwidth}{@{}lX@{}}
\toprule
\textbf{Domain} & \textbf{Example Questions} \\
\midrule
Dementia--Sport &
What proportion of studies are randomized controlled trials? \newline
How often is study design explicitly stated versus inferred? \newline
Are reporting guidelines (e.g., CONSORT, SPIRIT) referenced? \\
\addlinespace
NCD &
Which study designs dominate by disease area? \newline
Has the proportion of RCTs versus observational studies changed over time? \newline
How frequently do studies describe protocols rather than report completed trials? \\
\bottomrule
\end{tabularx}
\end{table}

\begin{table}[ht]
\centering
\caption{Cross-PICOS and research-waste–focused CRS competency questions}
\label{tab:crs_cross_picos}
\begin{tabularx}{\textwidth}{@{}lX@{}}
\toprule
\textbf{Domain} & \textbf{Representative Questions} \\
\midrule
Dementia--Sport &
What proportion of studies are fully PICOS compliant? \newline
Which PICOS elements most commonly drive exclusion decisions? \newline
Are there clusters of studies focused on reporting quality rather than clinical evidence? \\
\addlinespace
NCD &
Which NCD subdomains exhibit the highest PICOS-related research waste? \newline
Are there recurring patterns of partial compliance (e.g., P and I reported, C missing)? \newline
How often do studies duplicate existing evidence without extending populations or outcomes? \\
\bottomrule
\end{tabularx}
\end{table}

\begin{table}[ht]
\centering
\caption{High-level CRS reasoning modes used to address different query types}
\label{tab:crs_reasoning_modes}
\renewcommand{\arraystretch}{1.15}
\begin{tabularx}{\textwidth}{@{}lY@{}}
\toprule
\textbf{Reasoning Mode} & \textbf{Typical Question Characteristics} \\
\midrule
Structured querying & Counts, proportions, trends, and rankings derived from metadata or PICOS flags. \\
Semantic synthesis & Explanatory, interpretive, or narrative questions requiring full-text evidence. \\
Hybrid reasoning & Queries combining cohort definition with explanatory justification. \\
\bottomrule
\end{tabularx}
\end{table}

Together, these example questions illustrate how explicit PICOS formalization enables structured querying, explanation, and identification of research waste signals without requiring access to protected system outputs.

\end{document}